\documentclass[letterpaper, 10 pt, conference]{ieeeconf}  %

\IEEEoverridecommandlockouts                              %

\usepackage{amsmath}
\usepackage{mathtools}
\usepackage{booktabs}
\usepackage{lipsum}
\usepackage{float}
\usepackage{xspace}
\usepackage{cite}
\usepackage{graphicx}
\usepackage{wrapfig}
\graphicspath{{./figures/}}
\usepackage{caption}
\usepackage{subcaption}
\usepackage[inline]{asymptote}
\usepackage{tikz}

\makeatletter
\def\blfootnote{\xdef\@thefnmark{}\@footnotetext}
\makeatother
\usepackage{hyperref}

\newcommand{\gray}[1]{{\color{darkgray}{\textbf{#1}}}}

\begin{document}

\title{\LARGE \bf
RoRD: Rotation-Robust Descriptors and Orthographic Views for Local Feature Matching
}

\author{Udit Singh Parihar$^{*1}$,  Aniket Gujarathi$^{*1}$, Kinal Mehta$^{*1}$, Satyajit Tourani$^{*1}$, Sourav Garg$^2$, \\ Michael Milford$^2$ and K. Madhava Krishna$^1$ %

\thanks{$^{*}$Denotes authors with equal contribution}
\thanks{$^{1}$Robotics Research Center, IIIT Hyderabad.}
\thanks{$^{2}$QUT Centre for Robotics, Queensland University of Technology (QUT), Australia.}
\thanks{This research is supported by MathWorks.}
\thanks{\text{Code and dataset link}: \href{https://github.com/UditSinghParihar/RoRD}{https://github.com/UditSinghParihar/RoRD}}
\thanks{{In the IEEE's published version of this paper (DOI: 10.1109/IROS51168.2021.9636619), SIFT's results in the `Rotated' and `Average' columns of Table~\ref{mma_table} are incorrect, which we have now corrected in this arXiv version and over GitHub. This change only affects the performance ranking of MMA evaluation on the HPatches dataset, where our proposed method still ranks the best when considering standard and average settings. Other results in the paper which benchmark against SIFT, that is, Pose Estimation and Visual Place Recognition, are \textit{not} affected.}}
}

\maketitle
\global\csname @topnum\endcsname 0
\global\csname @botnum\endcsname 0
\thispagestyle{empty}
\pagestyle{empty}

\begin{abstract}

The use of local detectors and descriptors in typical computer vision pipelines works well until variations in viewpoint and appearance change become extreme. Past research in this area has typically focused on one of two approaches to this challenge: the use of projections into spaces more suitable for feature matching under extreme viewpoint changes, and attempting to learn features that are inherently more robust to viewpoint change. In this paper, we present a novel framework that combines the learning of invariant descriptors through data augmentation and orthographic viewpoint projection. We propose \textit{rotation-robust} local descriptors, learnt through training data augmentation based on rotation homographies, and a \textit{correspondence ensemble} technique that combines vanilla feature correspondences with those obtained through rotation-robust features. Using a range of benchmark datasets as well as contributing a new bespoke dataset for this research domain, we evaluate the effectiveness of the proposed approach on key tasks including pose estimation and visual place recognition. Our system outperforms a range of baseline and state-of-the-art techniques, including enabling higher levels of place recognition precision across opposing place viewpoints, and achieves practically useful performance levels even under extreme viewpoint changes. We reduce pose estimation error by $86.72 \%$ relative to state of the art.
\end{abstract}

\begin{figure}
  \includegraphics[width=\columnwidth]{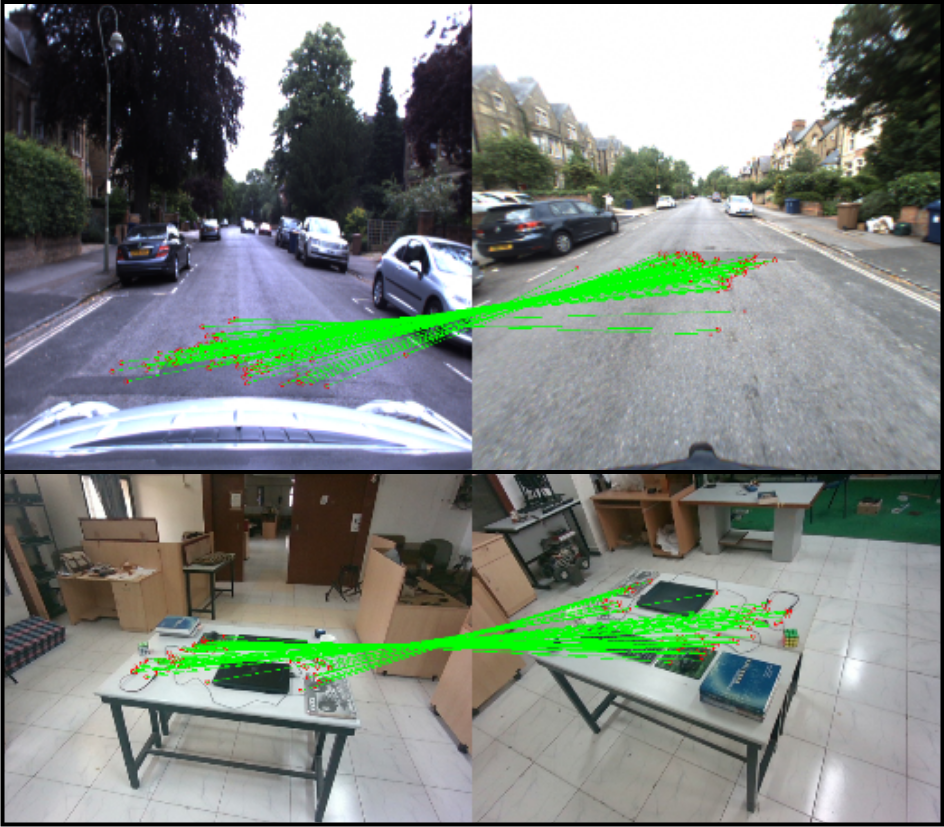}
  \caption{Local feature matches using RoRD. Our method RoRD finds precise local feature correspondences under extreme viewpoint ($\sim$180$^{\circ}$) changes for both indoor and outdoor sequences. }
\end{figure}
\section{INTRODUCTION}

\begin{figure*}
\includegraphics[width=\textwidth]{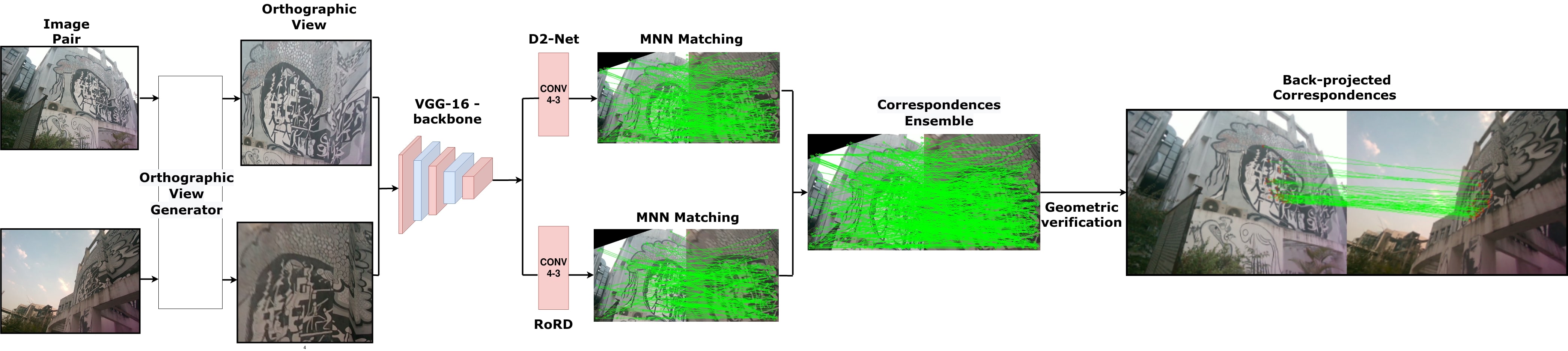}
\centering
\caption {RoRD pipeline.
Our approach takes a pair of perspective view images with significant viewpoint as input to the Orthographic View Generator, which aligns the camera to the plane-surface-normal to generate the top views, as explained in Section \ref{subsec:ortho}. These top views are then passed to an ensemble of \textit{Vanilla D2-Net}\cite{D2Net} and \textit{RoRD} techniques (ours). This approach creates precise feature correspondences that are robust to variations in viewpoints.}
\label{fig:pipeline}
\end{figure*}

The use of local detectors and descriptors is widespread in computer and robotic vision problems that include descriptor matching between an image pair\cite{superglue, D2Net, superpoint, s2d}, relative pose estimation\cite{s2d, superglue}, front-end pipelines for SLAM\cite{earlybird}, and image retrievals\cite{earlybird, RF}. Traditional local descriptors \cite{SIFT, SURF} typically employ a detector-descriptor pipeline; the detector comprises sparse or dense keypoints, while the descriptor characterizes a patch around the detector typically using a high dimensional vector. However, the performance of tasks like image retrieval, and pose estimation that rely on these classical descriptors degrades in the presence of large and disparate viewpoint changes.

The rising popularity of Deep Architectures has resulted in deep local detectors and descriptors \cite{superpoint, D2Net, r2d2}, which bypass the need to hand-craft features for the task at hand. Nonetheless, even popular implementations such as the SuperPoint Detector and Descriptor\cite{superpoint} and D2-Net \cite{D2Net} do not handle very large viewpoint changes.

In this paper we address the challenge of extreme viewpoint change by making the following contributions:
\begin{enumerate}
    \item We extend the operating range of deep descriptors such as D2-Net~\cite{D2Net} to function under extreme viewpoint change, by augmenting data with a succession of homographic transforms and learning \textit{rotation-robust} descriptors trained on this augmented data, achieving better generalization across both different environment types and perspective transformations;
    
    \item We present a new dual-headed D2-Net model with a \textit{correspondence ensemble} which combines vanilla and rotation-robust feature correspondences, leading to superior performance under all rotation variations;
    
    \item We introduce a new dataset that enables development and testing in this domain, \textit{DiverseView Dataset}, comprising 3 sequences of various outdoor and indoor images captured under extreme viewpoint variations;
    
    \item We evaluate the performance of these rotation-robust descriptors on a variety of tasks, datasets, and appropriate metrics, namely,
    \begin{enumerate}
        \item Descriptor Matching, evaluated using Mean Matching Accuracy (MMA) metric on the HPatches~\cite{HPatches} dataset, demonstrating improved Mutual Nearest Neighbor (MNN) based matches compared to~\cite{SIFT, superpoint, D2Net},
        \item Pose Estimation, evaluated by the relative angular error between actual and estimated poses on the \textit{DiverseView} dataset, demonstrating improved pose precision (orientation and translation) estimates compared to baselines~\cite{D2Net,superglue,SIFT}, 
        \item Visual Place Recognition (VPR), evaluated using the recall metric on the Oxford RobotCar dataset~\cite{RobotCarDatasetIJRR}, demonstrating a four fold improvement in opposite-view place matching percision (7 m) compared to the 30m achieved by previous state-of-the-art~\cite{LOST,SIFT,D2Net,superglue}, even under different camera intrinsics and illumination conditions. 
    \end{enumerate}

\end{enumerate}

\section{RELATED WORKS}

Here we briefly cover relevant research on local feature correspondences and orthographic view generation.

\subsection{Local Feature Correspondences}

Traditional hand-crafted feature methods such as SIFT\cite{SIFT}, SURF\cite{SURF}, ORB\cite{orb}, and BRISK\cite{brisk} have been successfully used in various computer vision applications for over a decade. However, these methods degrade significantly when exposed to extreme variations in scene appearance and camera viewpoint. More recently, the focus has shifted to Convolutional Neural Networks (CNNs) for the task of feature extraction, which offer promising advances in capability with respect to these challenges.

Recent research such as D2-Net\cite{D2Net} and SuperPoint\cite{superpoint} employ a CNN-based method to obtain both keypoints and descriptors. D2-Net\cite{D2Net} uses a detect-and-describe approach for sparse feature extraction, rather than performing early detection using the low-level features. This approach performs well or even better under challenging conditions such as illumination changes and weakly textured scenes. 
\textit{Sparse-to-Dense}\cite{s2d} uses a siamese architecture to learn its local descriptors. Similar to D2-Net\cite{D2Net}, it uses VGG-16 as its backbone.

LBP-HF~\cite{LHF} introduces a method to obtain rotation-invariant image descriptors based on~\cite{lbp}. Its rotation-invariant features are based on local binary pattern histograms. RI-LBD~\cite{lribp} employs a learning-based method to obtain local binary descriptors. It jointly learns the projection matrix and each pattern's orientation to obtain rotation-invariant local binary descriptors. Other CNN-based alternatives developed for rotation-equivariance \cite{gconv,e2cnn,cycnn} require re-designing of CNN architectures, whereas our proposed method achieves rotation-robustness by training with in-plane rotated images and performing correspondence ensemble.

\subsection{Orthographic View Generation}

Past research has shown that orthographic view generation improves performance in tasks such as image retrieval~\cite{earlybird}, and feature matching~\cite{RF}. \textit{Monolayout}~\cite{monolayout} predicts an amodal scene layout using an encoder to extract features at multiple scales. \textit{Rectified features}~\cite{RF} performs perspective unwarping via the help of 3D information. It uses~\cite{monodepth2} to first predict a dense depth map and eventually clusters the surface normals for each point.  After clustering, it then computes a rectifying homography to align the camera with the resulting surface normal. \cite{abbas2019geometric} estimates the homography that maps the perspective view to the bird's eye view by predicting vertical and horizontal vanishing lines using a CNN. \textit{Earlybird}~\cite{earlybird} performs visual place recognition for indoor environments by using floor features, estimating a fixed homography via the RANSAC + 4 point algorithm~\cite{mvg}.~\cite{rightangled} introduced a spatial transformer GAN for generating boosted Inverse Perspective Mapping (IPM), enabling improved detection of road markings and semantic interpretation of road scenes. Approaches like \cite{earlybird} and \cite{abbas2019geometric} obtain the necessary orthographic views through known a-priori information. In this work, we show that orthographic views alone are not sufficient when used with existing local features under significant viewpoint variations, and that rotation-robust descriptors are necessary to better leverage such views to achieve superior performance.

\section{Methodology}

The  main objective of this work is to determine accurate pixel-level correspondences  between  images  under  extreme  viewpoint variations. Here we introduce our proposed method for performing precise local feature matching by learning rotation robust descriptors invariant to large viewpoint changes.
We further extend the applicability of the proposed descriptors through a correspondence ensemble and orthographic view generation approach, as discussed in subsequent sections. An ensemble architecture allows us to have the illumination and scale invariance of the original D2Net, while RoRD makes the network robust to high viewpoint changes.

\subsection{Feature correspondences}
\label{feature-corres-method}

\begin{figure}
    \centering
    \includegraphics[height=100px]{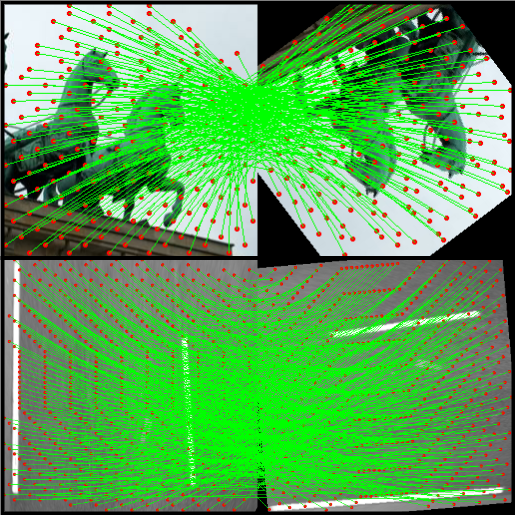}
    \caption{Example images of the training data used by RoRD. First row shows training pair for PhotoTourism Dataset and second row shows it for the road patches on the Oxford RobotCar Dataset.}
    \label{fig:train_pair}
\end{figure}

\subsubsection{Rotation-Robust Descriptors}
Given an input image {$I_1$}, we generate a corresponding image pair {$I_2$} using a uniformly distributed discrete random homography transformation consisting of $0^\circ-360^\circ$ rotations, as depicted in Figure~\ref{fig:train_pair}. This enables the generation of a large number of images with ground-truth pixel-level annotations in a self-supervised manner. Along with a series of in-plane rotations, we also incorporate perspectivity, scaling, and skewness transformations to further simulate realistic viewpoint changes. We use the PhotoTourism dataset for this training as detailed in the dataset section. 

Our approach uses VGG-16 as the base architecture, similar to D2-Net\cite{D2Net}, while fine-tuning the last layer and keeping the earlier layers' weights frozen. Input to the network is a cropped $400\times400\times3$ image region and a random rotation homography $H_{R}(\theta)$ is used to rotate the image in-plane by $\theta$ degrees to create the training pair. We follow the same training procedure of \textit{detect} and \textit{describe} as defined in~\cite{D2Net}, with primary difference being the type of training data (and augmentation) used which in our case is based on in-plane rotations. The local descriptors thus obtained through this training procedure are referred to as \textit{RoRD} (\textit{Ro}tation \textit{R}obust \textit{D}escriptors).

\subsubsection{Correspondence Ensemble}
While the above procedure leads to significant performance gains in local feature matching under extreme viewpoint variations, we observe that catastrophic forgetting could occur when viewpoint variations are rather moderate or none. To overcome this limitation, we consider two solutions: one based on training a single model and other based on training two separate heads of the same model. Note that the first solution is merely chosen as a baseline to justify the use of the second solution, as later demonstrated in Section~\ref{section:results}. For the single model training, referred to as RoRD-Combined, we use the combined training data from 3D depth and pose information (as was used in the original D2-Net) and in-plane rotation and random homographies (as proposed in this work). 

For the second solution, we propose a Correspondence Ensemble (CE) method where we use a dual-headed D2Net as shown in Fig.~\ref{fig:pipeline}. The first head corresponds to the output of vanilla D2-Net (trained with data as per the original method), while the second head corresponds to RoRD (trained using in-plane rotations data). This dual-headed approach improves model efficiency due to a common backbone shared across both the models since we only train the final layer. Once we have keypoints and descriptors from the two heads, mutual nearest-neighbor based feature correspondences are obtained between an image pair independently for each head. Using the descriptor distances of the combined set of correspondences obtained from the two heads, we select the top $50\%$ of the correspondences which have the lowest distance values. These correspondences are then filtered by RANSAC-based geometric verification to obtain a final set of feature correspondences.

\begin{figure}[h!]
\includegraphics[width=0.4\textwidth]{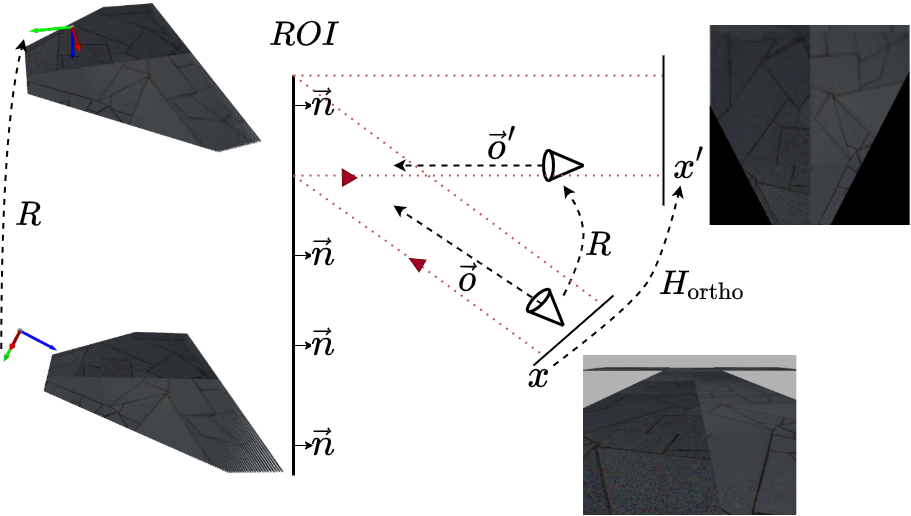}
\centering
\caption{Orthographic View Generation. We use 3D geometry to obtain our necessary \textit{ROI} patches in the form of orthographic projection. Depth information is used to generate surface normals of our desired \textit{ROI}. The camera is aligned such that its optical axis is antiparallel to the surface normal and obtains the desired orthographic view.}
\label{fig:surfacenormalfig}
\end{figure}

\subsection{Orthographic view generation} \label{subsec:ortho}

Most existing \cite{superglue, superpoint, D2Net} feature matching methodologies use perspective view images for detection, description and matching of high-confidence keypoints. These methods work reliably on image pairs with smaller changes in viewpoint or rotation but degrade as camera viewpoint becomes more extreme. As previously demonstrated in~\cite{RF,earlybird}, orthographic views can help improve local feature matching due to an increased visual overlap through such projections. However, such views alone may not be sufficient to achieve this goal. In this work, we show that rotation-robust features are additionally required to better leverage orthographic views, especially when dealing with extreme viewpoint variations. Here, we describe our approach to generating orthographic views.

Our orthographic view generation process utilizes 3D information and has two possible operating modes. In scenes where the layout is varied and the camera is moving in 6-DoF, it utilizes ROI planar regions and surface normals. In relatively consistent scene layout scenarios like self-driving cars, we use a fast, lightweight IPM based approach to obtain top views of the road.

\subsubsection{Surface normal based orthographic view generation}
\label{ortho-surfnormal}
For the DiverseView dataset, we use 3D geometry to obtain orthographic views as shown in Fig.~\ref{fig:surfacenormalfig}. 
Given a dominant planar region, we use depth information (obtained through Intel Realsense D455 camera) to generate the 3D point cloud of the desired \textit{ROI} and calculate the surface normal of this plane using~\cite{open3d}. Further, we align a virtual camera such that its optical axis becomes antiparallel to the surface normal of the plane and its location lies directly above the centroid of the point cloud. Images taken using the virtual camera location and the original camera location are related by a rectifying homography, $H_{ortho}$. This matrix enables us to obtain the desired orthographic view $x ^{'}$ from the perspective view $x$.

We use the rotation matrix $R$ to align the camera optical axis, $\Vec{o}$, with the semantic-plane-specific surface normal, $\Vec{n}$, having distance $d$ apart from the camera. To account for the translation between the perspective view and the orthographic view, we also incorporate a fixed translation $\Vec{t}$ to ensure that our camera is placed at the right location.

\begin{align}
    \label{eq:rmatrix}
\Vec{v} &= \Vec{o} \times \Vec{n} \\
R &= I + [\Vec{v}]_{x} + [\Vec{v}]_{x}^{2}\times\frac{1-\Vec{o} \cdot \Vec{n}}{\lVert \Vec{o} \times \Vec{n} \rVert^{2}}
\end{align}

$\Vec{v}$ is the vector perpendicular to $\Vec{o}$ and $\Vec{n}$ and $[v]_{x}$ is its skew-symmetric matrix. We then use this rotation matrix to obtain the rectifying homography matrix, $H_{ortho}$, which enables us to obtain the desired orthographic view. 
\begin{align}
    \label{eq:homography}
x^{'} &= K \: (R - \frac{\Vec{t} \Vec{n}^T}{d} ) \: K^{-1} \: x  \\
x^{'} &= H_{ortho} \: x
\end{align} 

$K$ is the intrinsic matrix and $K (R - \frac{\Vec{t} \Vec{n}^T}{d} ) K^{-1}$ corresponds to the rectifying homography matrix, $H_{ortho}$.

\subsubsection{Inverse Perspective Mapping (IPM) based orthographic view generation}
\label{IPM}
Considering an image pair in autonomous driving scenarios, with high-viewpoint change and relative translation, the overlapping region between the pair consists largely of the road. While seemingly aliased, in practice, roads contain many discriminative features. We leverage the cracks and deformities on the road surfaces to generate deep features which are robust to $180^0$ rotations, different intrinsics, and illumination change.      
We focus on the benchmark Oxford RobotCar Dataset~\cite{RobotCarDatasetIJRR}. Due to the fixed camera geometry with respect to the road, a single homography can rectify the front camera image, and similarly, another homography can rectify the rear camera image. These homographies are generated by mapping parallel lane markers on the road to an appropriate sized polygonal region, and the best fit homography is calculated.

As shown in Table~\ref{rt_table} and \ref{vpr_local}, both the aforementioned methods boost the performance of local feature matching for their respective datasets.

\section{EXPERIMENTAL SETUP}

In this section we describe the key benchmarks and newly contributed datasets used to evaluate performance, as well as the range of evaluation metrics used.

\subsection{Datasets}
\label{datasets}
We report our results on two standard benchmark datasets and the new DiverseView dataset.

\subsubsection{PhotoTourism \cite{Phototourism}} This dataset was only used for training. We used the following seven scenes for this purpose: the Brandenburg Gate, Buckingham Palace, Grand Place Brussels, Hagia Sophia Interior, Taj Mahal, Temple Nara Japan and Westminster Abbey.

\subsubsection{HPatches \cite{HPatches}} We use this dataset for evaluating local feature matching. For this purpose, we use the original dataset as \textit{Standard HPatches} (108 out of 116 scenes) as done in~\cite{D2Net}. We additionally proposed a variant which based on random in-plane rotation of images from 0 to $360^{0}$, referred to as \textit{Rotated HPatches}.

\subsubsection{Oxford RobotCar Dataset \cite{RobotCarDataset}} We evaluate the Visual Place Recognition (VPR) performance enabled by our underlying methods using two different traversals of the benchmark Oxford RobotCar Dataset: Training Route ($2014/07/14-14:49:50$) and Testing Route ($2014/06/26-09:24:58$). Our training data comprises $6200$ images obtained from both the front (\textit{Bumblebee}) and rear (\textit{Grasshopper}) facing cameras. Using the same self-supervised training protocol as described in Section~\ref{feature-corres-method} and orthographic view generation process described in Section~\ref{subsec:ortho}, top-views of images are used and training image pairs are formed by in-plane rotations. Note that front image is never compared to the rear image during training, and training and test data belong to geographically distinct locations.

\subsubsection{DiverseView dataset} 
Due to the lack of publicly available datasets with high viewpoint changes and camera rotations, we introduce a new dataset comprising both outdoor and indoor scenarios. The DiverseView dataset consists of 3 scenes of indoor and outdoor locations, with images captured at high viewpoint changes and camera rotations (up to 180$^{\circ}$). For the data collection, we have used the Intel RealSense D455 camera with RGB, Depth, and IMU sensors. The DiverseView dataset along with all sensor information for ground truth and orthographic view generation purposes will be released publicly. 

The three sequences of this dataset are as follows: \textit{Sequence 1} consists of $1138$ images of an office-desk scene captured by moving around it in a semi-circular path, thus exhibiting extreme rotation variations. \textit{Sequence 2} consists of $2534$ images obtained from a $360^{\circ}$ traversal about a table-top. \textit{Sequence 3} consists of $3931$ images captured around a building with graffiti-art with varied camera viewpoints and rotations in low-lighting dusk conditions. For this dataset, evaluation is done using the model trained on the PhotoTourism dataset, thus showcasing the generalizability of proposed RoRD.

\subsection{Evaluation}
\label{sec:eval-metric}
The versatility of our method is evaluated by showing results using three different metrics on three different tasks. For the first task, we evaluate the performance through an NN-based matching using MMA, as was done in prior work~\cite{D2Net}, on the HPatches dataset. We then evaluate our methods and the baselines on the task of Pose Estimation, as performed in~\cite{superglue}, on the \textit{DiverseView} dataset. Finally, we evaluate the quality of local feature matches on the Oxford RobotCar dataset for opposing-viewpoint visual place recognition.

\subsubsection{Local feature matching using MMA}
We follow the standard protocol~\cite{D2Net} for evaluating nearest neighbor feature matching on HPatches dataset. A match is considered to be correct if the backprojection of the keypoint using the ground truth homography matrix falls within a specific pixel threshold. Computing the mean matching accuracy (MMA) of all pairs, over varying thresholds, all the models are then compared on the two variants of HPatches - Standard HPatches and Rotated HPatches. We use the following baseline comparison methods: \textit{D2-Net}~\cite{D2Net} and \textit{SuperPoint}~\cite{superpoint}, using the trained model provided by the respective original authors; \textit{SIFT}~\cite{SIFT}, using its OpenCV implementation. From the proposed methods, we include RoRD, RoRD + CE (Correspondence Ensemble) and RoRD-Combined, as described in Section~\ref{feature-corres-method}.

\subsubsection{Pose Estimation using Rotation and Translation error}
We use DiverseView dataset for this task. To generate precise ground truth poses for the relative pose (R/t) estimates, we used RTABMAP SLAM~\cite{rtabmap} with sensor fusion. Relative angular error, $|| \; log(\hat{R}R_{gt}^{T}) \; ||/\sqrt{2}$ in degrees is used to calculate the error between the predicted, $\hat{R}$ and ground-truth $R_{gt}$ rotations. A logarithmic mapping converts the difference rotation matrix, $\hat{R}R_{gt}^{T}$, to an axis-angle representation in degrees. The Euclidean norm is used between the ground-truth translation and predicted translation. For evaluation, we sample $10$ representative images across the individual dataset sequences. Each representative image is then matched with $100$ images evenly distributed across the dataset sequence and then the extracted $R/t$ is compared against the ground-truth. Finally, $10\times100$ $R/t$ residual errors are averaged to obtain final error for that dataset. We compare our method against D2-Net~\cite{D2Net}, SuperPoint+SuperGlue~\cite{superglue} and SIFT~\cite{SIFT} on both perspective and orthographic views in R/t estimation.

\subsubsection{Visual Place Recognition using Local Feature Matching}
We evaluate the effectiveness of using \textit{rotation robust features} and \textit{orthographic images} by performing \textit{visual place recognition} on the Oxford RobotCar dataset using GPS based priors~\cite{vysotska2015efficient}. $835$ front camera images are used as queries and each query image is compared against all the rear images (considered as reference database) within a $52$ m radius using GPS. The reference image having the maximum number of inlier correspondences is retrieved. We consider a match to be correct if the retrieved match lies within a $7$ m localization radius of the car's reverse traversal (visual overlap).

\begin{table}
\caption{Quantitative results for MMA on the HPatches dataset using pixel thresholds 6/8/10. We highlight the \textbf{\underline{first}}, \textbf{second} and \underline{third} best MMA values. We also provide averages over the results obtained by the Standard and Rotated HPatches dataset. Our Ensemble method RoRD + CE outperforms all other methods with RoRD and SIFT\textbf{\LARGE{*}} being the second and the third best performer for the averaged HPatches datasets, respectively. 
}
\label{mma_table}
\begin{center}
\resizebox{\columnwidth}{!}{
\begin{tabular}{l c c c}
\toprule
\textbf{Model} & \textbf{Standard} & \textbf{Rotated} & \textbf{Average}\\
\bottomrule
SIFT\cite{SIFT}\textbf{{*}} & 0.52/0.54/0.54 & \textbf{\underline{0.51/0.51/0.52}} & \underline{0.52/0.53/0.53} \\ 
SuperPoint\cite{superpoint} & 0.69/0.71/0.73 & 0.21/0.22/0.22 & 0.45/0.46/0.48 \\ 
D2-Net\cite{D2Net} & \textbf{0.73/0.81/0.84} & 0.17/0.20/0.22 & 0.45/0.50/0.53  \\ 
(Ours) RoRD & 0.68/0.75/0.78 & \underline{0.46/0.57/0.62} & 0.57/\textbf{0.66/0.70}  \\
(Ours) RoRD Comb. & \underline{0.71/0.78/0.81} &  0.44/0.54/0.59 & \textbf{0.58/0.66/0.70}  \\ 
(Ours) RoRD + CE & \textbf{\underline{0.79/0.84/0.86}} & \textbf{0.48/0.59/0.64} & \textbf{\underline{0.64/0.72/0.75}}  \\ \bottomrule

\end{tabular}}
\end{center}
\end{table}

\section{RESULTS}
\label{section:results}
In this section, we show the performance comparisons of different models on various tasks as explained in Section~\ref{sec:eval-metric}.

\subsection{Local Feature Matching} The quantitative performance of local feature matching on the HPatches dataset is presented in Table~\ref{mma_table} for three settings: standard, rotated, and the average of two. As mentioned in Section~\ref{sec:eval-metric}, we follow the standard protocol of~\cite{D2Net} and use mutual NN matching, evaluating the results using MMA for pixel thresholds 6, 8, and 10. Our proposed method \textit{RoRD + CE} (correspondence ensemble) consistently outperforms other methods including SIFT~\cite{SIFT}, ORB~\cite{orb} and SuperPoint~\cite{superpoint}.

For the Rotated HPatches dataset, SIFT achieves the best results\textbf{\LARGE{*}}\blfootnote{\textcolor{darkgray}{\textbf{\LARGE{*}}\normalsize{In the IEEE's published version of this paper (DOI: 10.1109/IROS51168.2021.9636619), SIFT's results in the `Rotated' and `Average' columns of Table~\ref{mma_table} are incorrect, which we have now corrected in this arXiv version and over GitHub: https://github.com/UditSinghParihar/RoRD. This change only affects the performance ranking of MMA evaluation on the HPatches dataset, where our proposed method still ranks the best when considering standard and average settings. Other results in the paper which benchmark against SIFT, that is, Pose Estimation and Visual Place Recognition, are \textit{not} affected.}}} and RoRD + CE achieves the second best, closely followed by RoRD. Overall, considering the average performance across standard and rotated HPatches, our ensemble method RoRD + CE achieves the best performance, while also outperforming RoRD-Combined which shows the effectiveness of dual-headed D2Net and correspondence ensemble over the combined data based training of a single model.

Since SuperGlue~\cite{superglue} is a matcher with inherent outlier rejection based on scene geometry, we have not included it in Table~\ref{mma_table} where no geometric verification (via RANSAC) is performed. For a fair comparison against SuperGlue, we consider RoRD+RANSAC which achieves an average MMA of 0.78/0.95 using a pixel threshold of 6/10, as compared to 0.71/0.73 achieved by SuperGlue.

\subsection{Pose estimation} 
\label{sec:result_pose_estimation}

Table~\ref{rt_table} shows results on the DiverseView dataset for rotation and translation estimation, relevant for downstream tasks like point cloud registration or 6-DoF localization. To evaluate the accuracy of the rotation and translation estimates, we follow the evaluation metric as described in~Section~\ref{sec:eval-metric}. In this evaluation, we consider two additional aspects on top of the previous analysis on HPatches: we use RANSAC-based geometric verification, and employ orthographic views. We show that our proposed methods (\textit{RoRD} and \textit{RoRD + CE}), although trained on the PhotoTourism dataset, can generalize well when inferring on the DiverseView dataset. This suggests that our methods are robust to changes in the data distribution due to changes in environment type and characteristics.
    
\begin{table}
\label{rt-table}
\caption{Quantitative results for the Pose Estimation(R/t) task on the DiverseView dataset. \textbf{First} and \gray{Second} best values are highlighted. For the perspective view setting, on sequences 1 and 3, \textit{RoRD} performs the best with our ensemble method \textit{RoRD + CE} being the second best. On sequence 2, \textit{SIFT} obtains the best orientation (rotation) estimates with \textit{RoRD} performing the second best. For the translation estimates, \textit{RoRD} and \textit{RoRD + CE} perform the best and second best respectively. For the orthographic view setting, on sequences 1 and 2, \textit{RoRD} performs the best with \textit{RoRD + CE} being the second best. For sequence 3, \textit{SIFT} and \textit{RoRD} perform the best and the second best.}
\label{rt_table}
\begin{center}
\resizebox{\columnwidth}{!}{%
\begin{tabular}{l c c c}
\toprule
\textbf{Model} & \textbf{Sequence 1} & \textbf{Sequence 2} & \textbf{Sequence 3}\\
\bottomrule
\multicolumn{4}{l}{\textbf{Perspective View}} \\
\bottomrule
SuperPoint+SuperGlue\cite{superglue} & 58.71/1.18 & 79.03/1.47 & 49.27/4.45 \\ 
D2-Net\cite{D2Net} & 68.76/1.02 & 80.28/1.44 & 48.33/4.55 \\ 
SIFT\cite{SIFT} & 48.43/1.12 & \textbf{59.81}/1.30 & 12.74/1.47 \\ 
(Ours) RoRD & \textbf{18.12/0.42} & \gray{61.90}/\textbf{1.15} & \textbf{10.17/1.38} \\ 
(Ours) RoRD + CE & \gray{20.69/0.44} & 65.55/\gray{1.21} & \gray{10.37/1.41} \\ 
\bottomrule
\multicolumn{4}{l}{\textbf{Proposed Orthographic View (OV)}} \\
\bottomrule
SuperPoint+SuperGlue\cite{superglue} & 54.73/0.80 & 77.85/1.13 & 44.61/4.37 \\ 
D2-Net\cite{D2Net}  & 63.69/0.87  & 80.71/1.29  & 51.04/4.80 \\ 
SIFT\cite{SIFT} & 13.04/0.26 & 17.84/0.34 & \textbf{5.11/0.76} \\ 
(Ours) RoRD  & \textbf{7.71/0.18} & \textbf{8.58/0.20} & \gray{7.79/1.03}
 \\ 
(Ours) RoRD + CE  & \gray{8.66/0.18}  & \gray{16.32/0.32} & 8.33/1.07 \\ \bottomrule

\end{tabular}}
\end{center}
\end{table}

\subsubsection{Perspective View}
\label{sec:result_perspective_view}
From Table~\ref{rt_table}, we observe that with perspective images as input to match local features for challenging viewpoints, for sequences \textit{1 and 3} \textit{RoRD} performs the best with our ensemble method \textit{RoRD + CE} being a close second. For sequence 2, we observe that SIFT~\cite{SIFT} can obtain the best rotation estimates with RoRD obtaining the best translation estimate. RoRD and our ensemble method RoRD + CE obtain second-best estimates for rotation and translation respectively. We notice that across all sequences, our method either performs best or second-best for both rotation and translation estimates. This bolsters the fact that for both indoor and outdoor sequences, to match local features for challenging viewpoints we need features that are invariant to rotation.
    
\subsubsection{Orthographic View}
\label{sec:result_planercnn}
We further improve upon the aforementioned results by using orthographic views instead of perspective views. We attribute this to the fact that when using orthographic views, the visual overlap increases and makes the task of feature matching easier, supporting the findings of~\cite{RF, earlybird}. However, here we also demonstrate that \textit{the orthographic view alone is not sufficient} and rotation-robust descriptors are also required in combination to be able to fully leverage such planar views based on performance enhancement. This is evident from Table~\ref{rt_table} that orthographic views lead to proportionally and consistently larger performance gains for RoRD than for other existing local feature methods. In Table~\ref{rt_table}, for sequences \textit{1} and \textit{2}, our method \textit{RoRD} outperforms other baselines. For sequence 3, we observe that \textit{SIFT} has the best performance with \textit{RoRD} being the second-best. We notice in the absence of rotation-invariant specific training regime, deep methods such as Superglue~\cite{superglue} and D2-Net~\cite{D2Net} struggle to consistently find accurate feature correspondences in such extreme cases. On average, across all datasets and both types of views, our proposed rotation-robust descriptors achieve superior performance.

For the DiverseView dataset, the dominant planes were assumed to be available through manual annotation (to be released along with the the dataset). Our initial investigation revealed that most of the existing methods suitable for planes and surface normal extraction such as PlaneRCNN~\cite{planercnn}, Multi-task RefineNet~\cite{nekrasov2019real} and DenseDepth~\cite{densedepth} catastrophically failed, which can be attributed to their lack of generalization or untypical viewpoints. This shows that image based 3D scene understanding methods are far from perfect. Nevertheless, we evaluated PlaneRCNN based dominant plane extraction for Sequence~3 and found that for the cases where plane extraction worked correctly, orthographic-view based pose estimation performed similar to that presented in Table~\ref{rt_table}.

\newcommand{\scaleOne}{0.07}

\begin{figure*}

\centering
\begin{tabular}{c c c}

 \includegraphics[scale=\scaleOne]{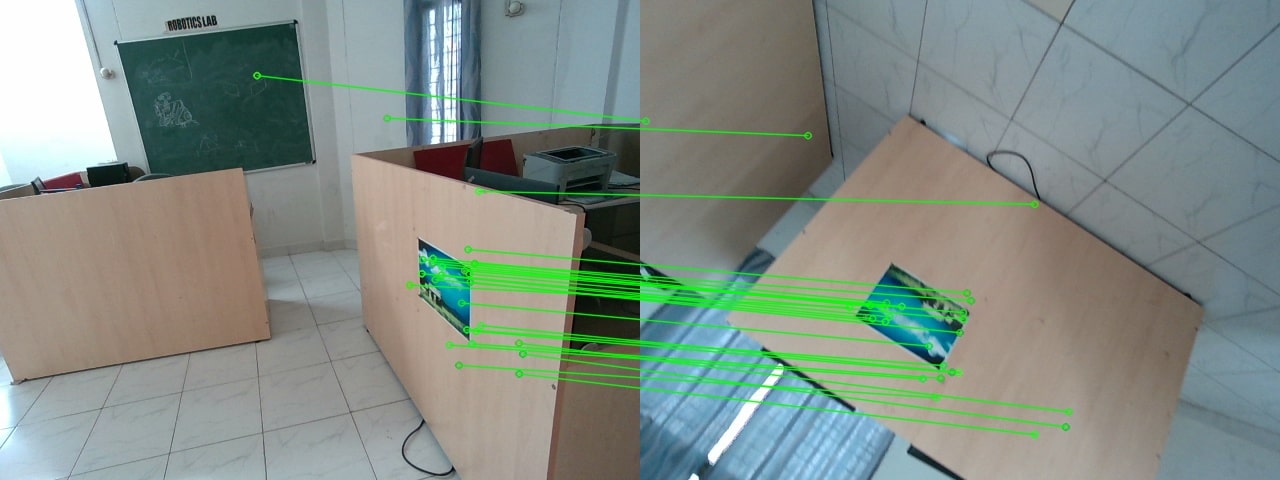}      & 
                          \includegraphics[scale=\scaleOne]{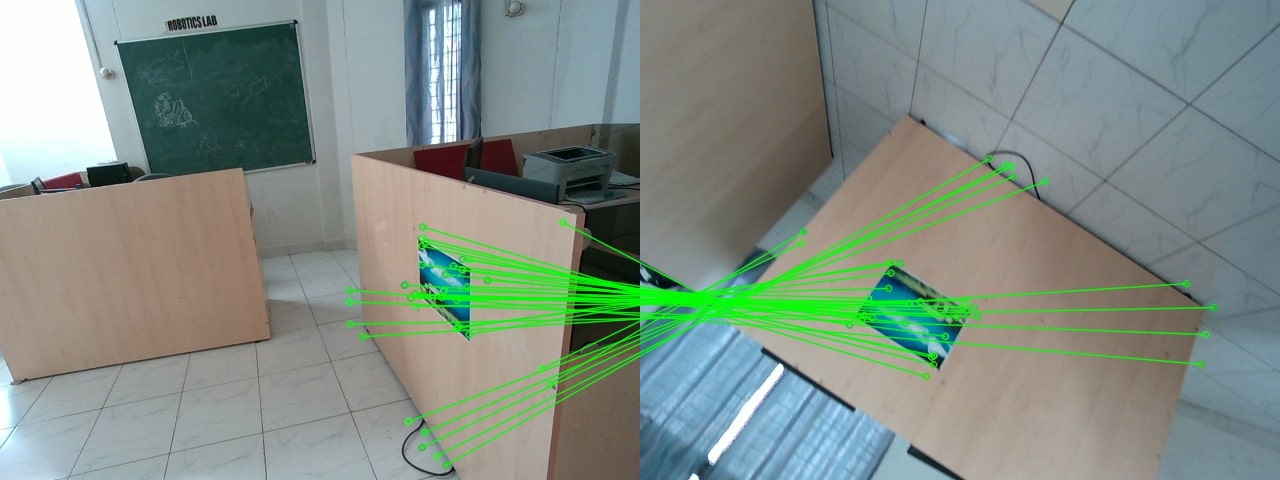} &
                          \includegraphics[scale=\scaleOne]{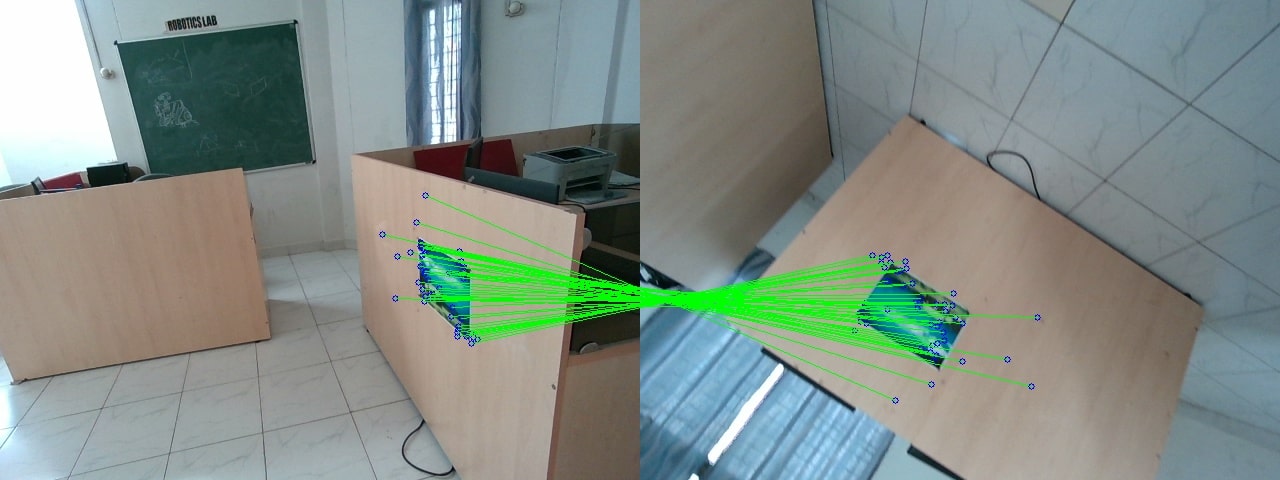}  \\
                          
\hline
 \includegraphics[scale=\scaleOne]{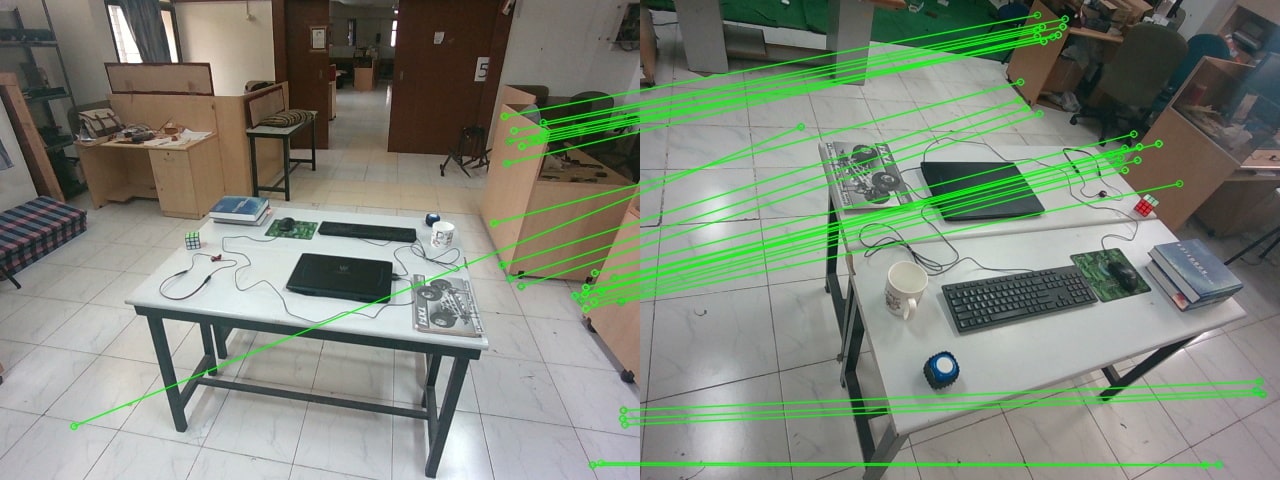}&
                          \includegraphics[scale=\scaleOne]{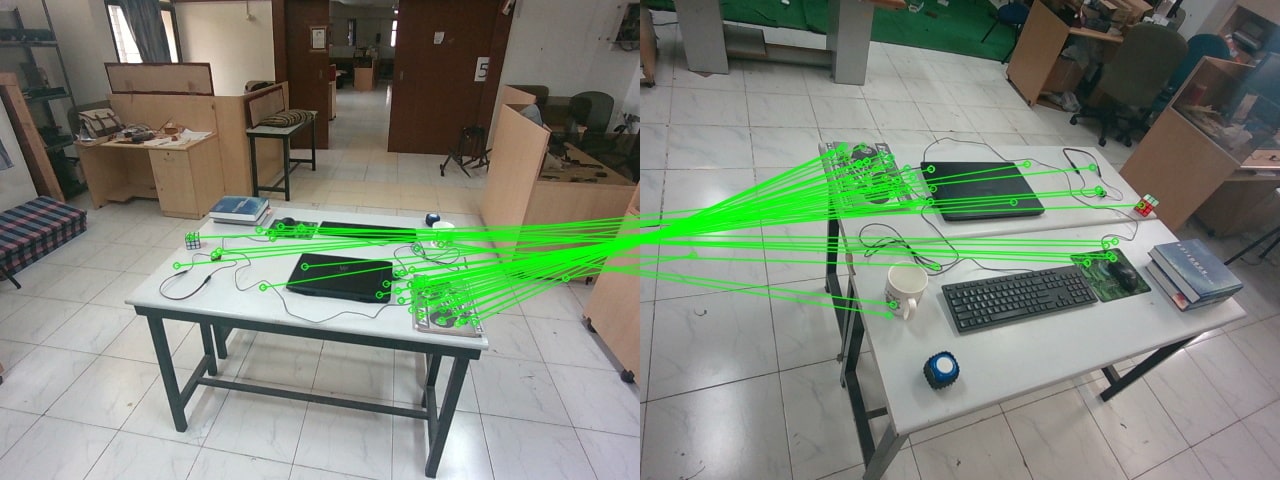} &
                          \includegraphics[scale=\scaleOne]{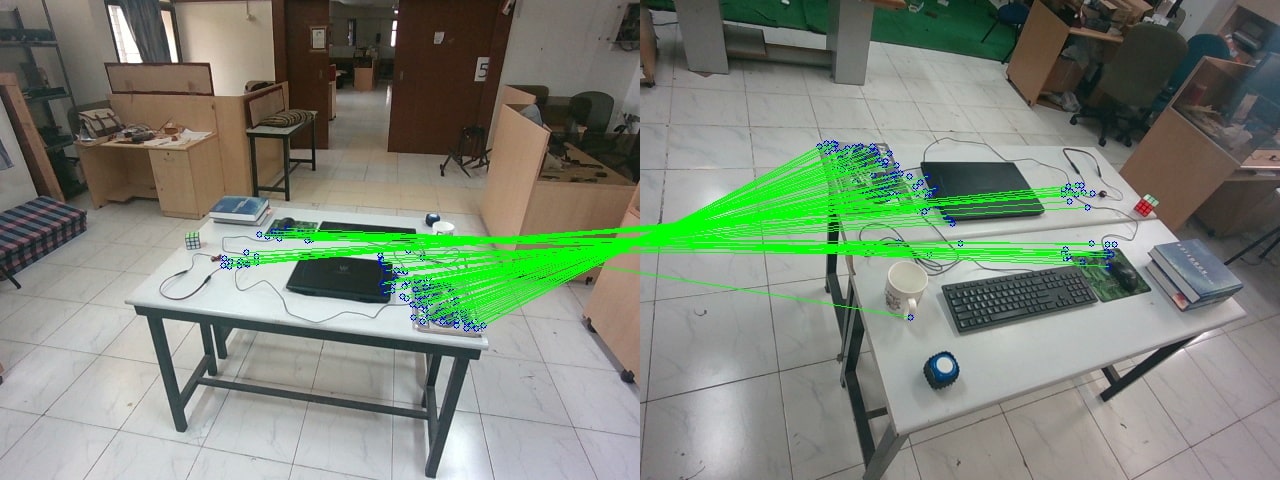} \\
\hline

 \includegraphics[scale=\scaleOne]{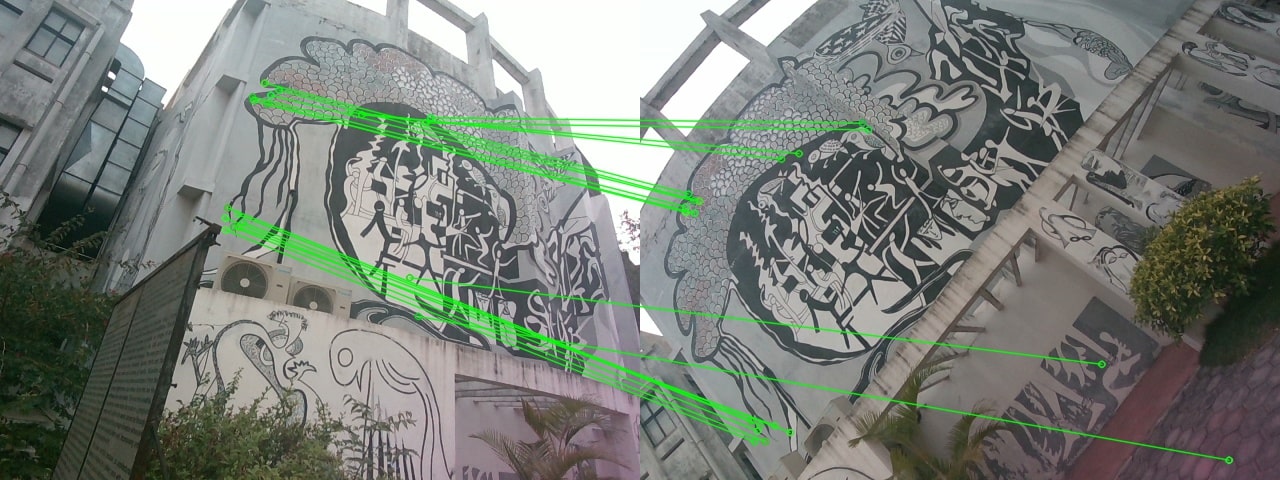}    & 
                          \includegraphics[scale=\scaleOne]{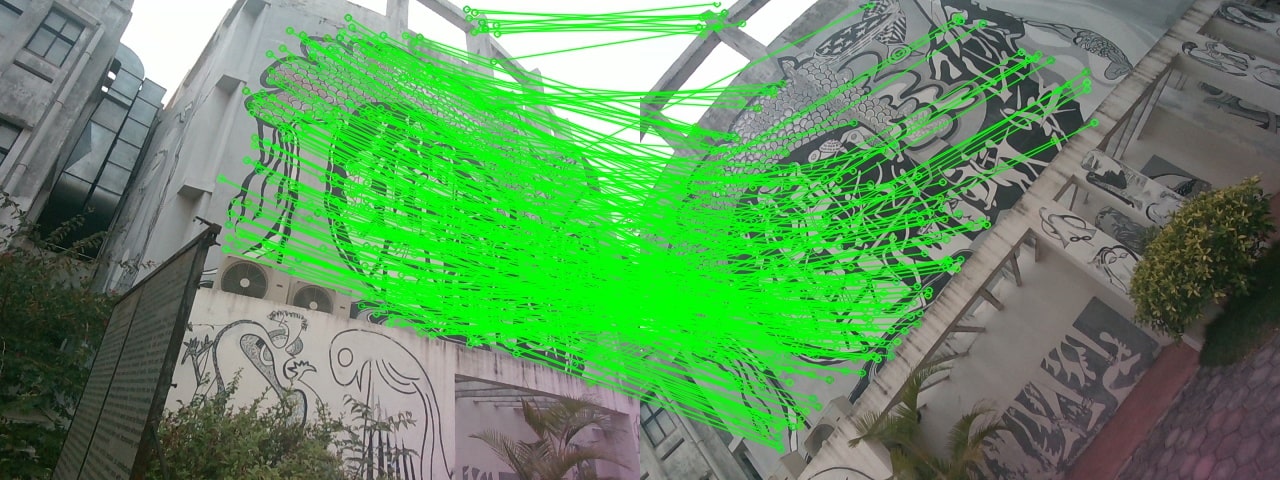} &      
                           \includegraphics[scale=\scaleOne]{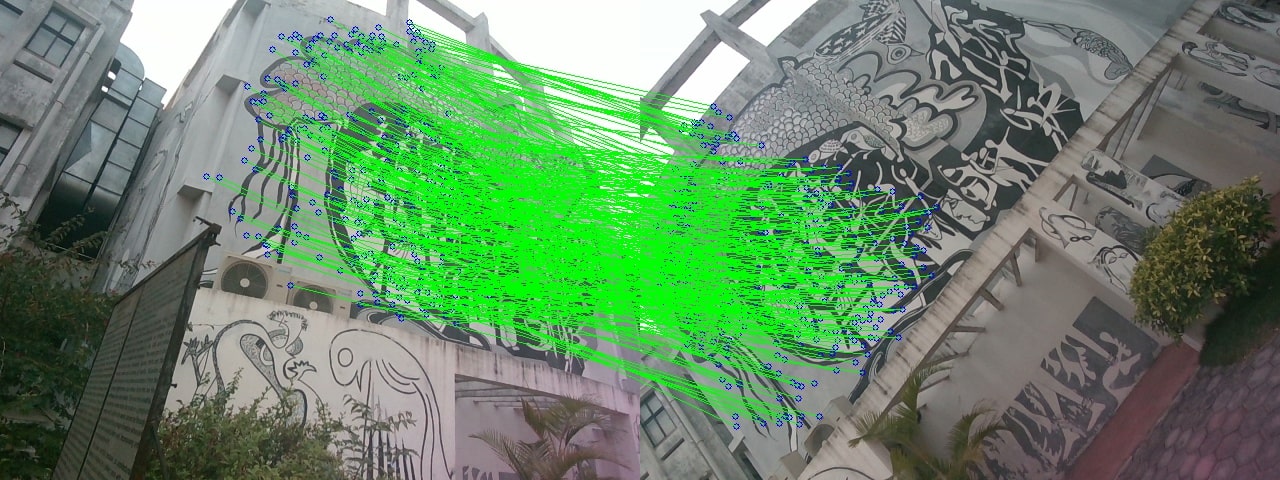}  \\
                           \hline
\small{D2-Net} &  \small{RoRD}  &   \small{RoRD + OV} \\ 
                    
\end{tabular} 

\caption{Qualitative results from the DiverseView dataset. Each row corresponds to a different sequence from the dataset. The top, middle and the bottom rows correspond to Sequence-1, Sequence-2 and Sequence 3 respectively. \textit{RoRD} leverages its learnt \textit{rotation-robust features} to obtain precise feature correspondences, outperforming vanilla D2-net\cite{D2Net}. Incorporating orthographic views with RoRD, \textit{(RoRD + OV)} further improves performance, and outperforms both D2-Net and RoRD for all three sequences.}
\label{fig:dv-qual}
\end{figure*}

\newcommand{\scaleTwo}{0.12}
\begin{figure}

\centering
\begin{tabular}{c}

 \includegraphics[scale=\scaleTwo, height=65px]{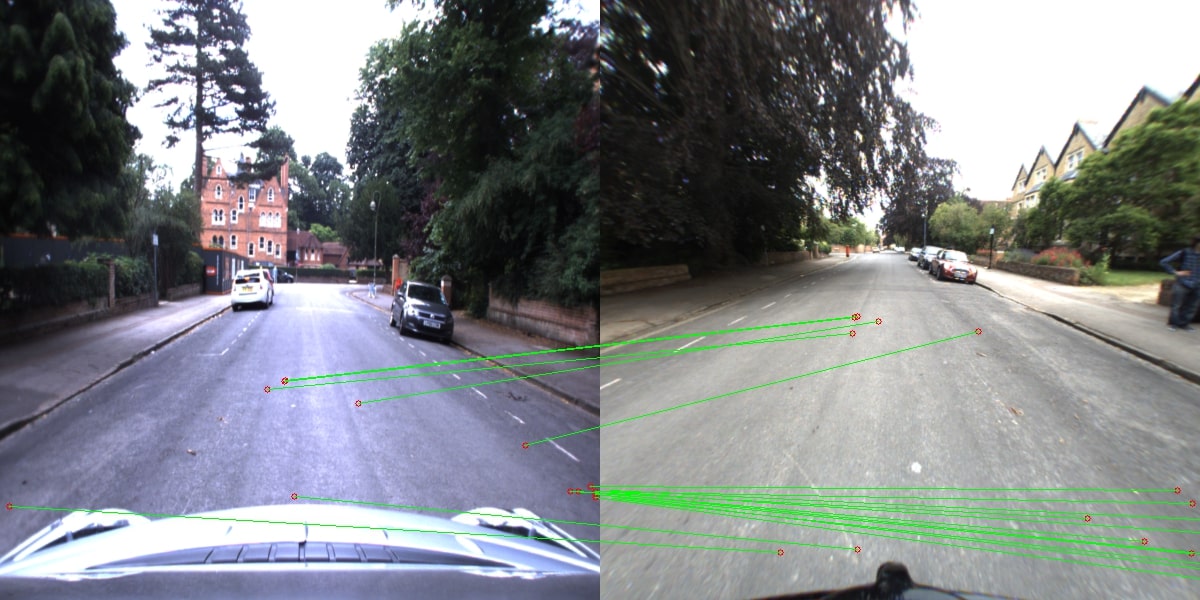}  \\    
\includegraphics[scale=\scaleTwo, height=65px]{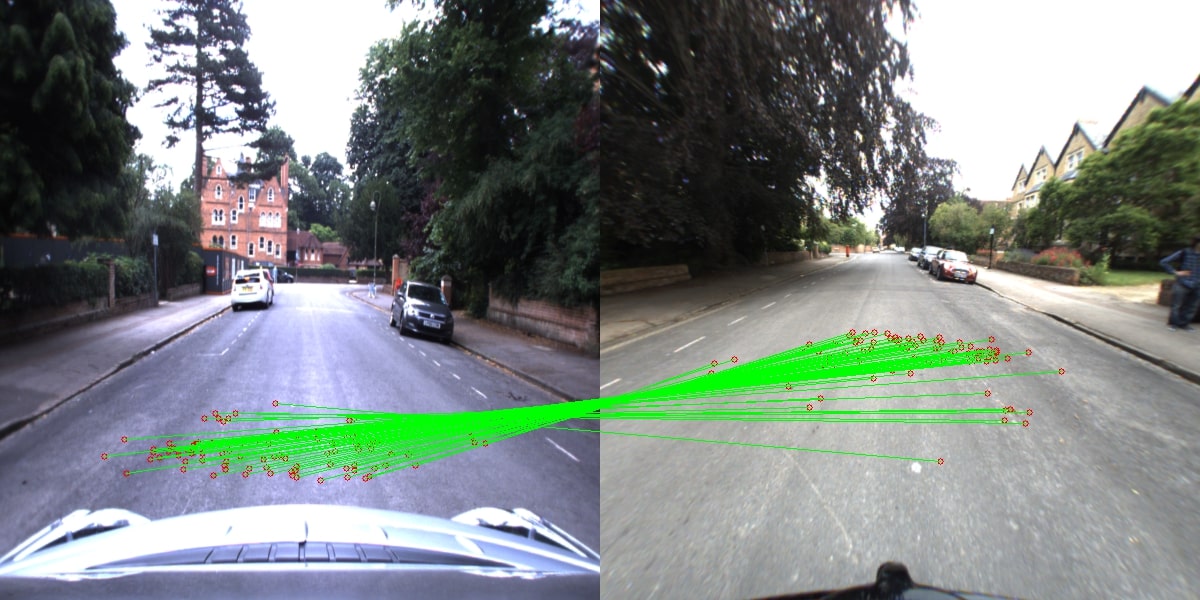}   \\ 
                    
\end{tabular} 

\caption{Qualitative results on the Oxford Robotcar Dataset. We show the feature correspondence results for \textit{D2-Net} (top) and \textit{RoRD} (bottom) when operating on orthographic views (In these figures we backproject the obtained correspondences to the perspective view). \textit{RoRD + OV} outperforms \textit{D2-Net + OV}.}
\label{fig:dv-qual2}
\end{figure}

\begin{figure}[h]

\centering
\begin{tabular}{c}

 \includegraphics[scale=\scaleTwo]{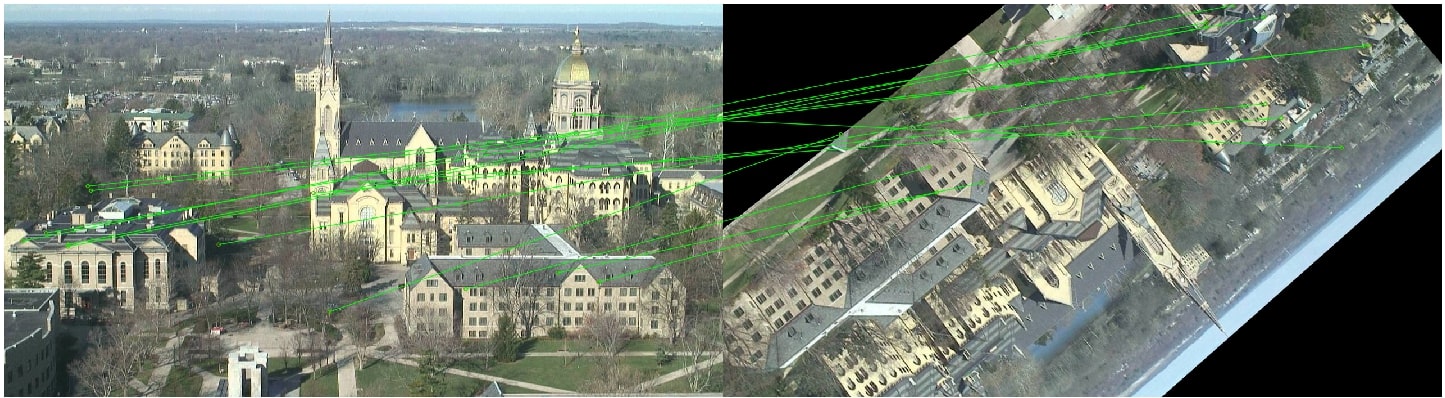}     \\
 \includegraphics[scale=\scaleTwo]{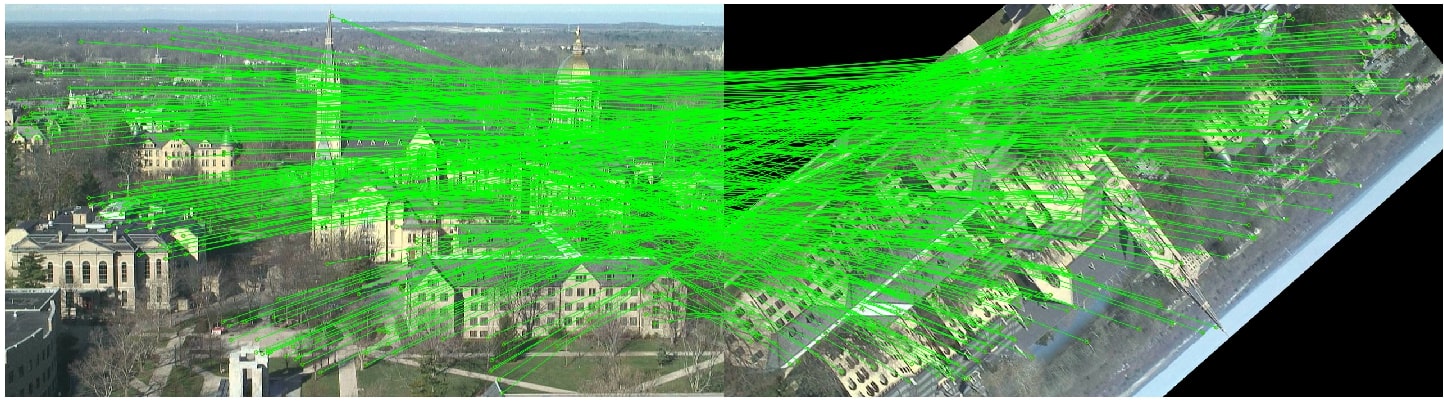}  \\ 
                    
\end{tabular} 
\caption{Qualitative results on the HPatches Dataset. Our method \textit{RoRD} (bottom) outperforms \textit{D2-Net} (top) by a significant margin.}
\label{fig:dv-qual3}
\end{figure}

\subsection{Visual Place recognition using local feature matching}
\label{section:results-robocar}

    Table~\ref{vpr_local} shows the quantitative results for visual place recognition using local feature matching for forward and rear view sequences \textit{(Opposite viewpoints)} on the Oxford RobotCar Dataset, using orthographic views. We observe that our method \textit{RoRD} has the best performance as compared to the baselines D2-Net~\cite{D2Net}, SuperPoint+SuperGlue (SP+SG)~\cite{superglue}, SIFT (Rectified Features)~\cite{RF}, and LoST~\cite{LOST}. LoST is the only perspective-view based global descriptor method included in this study for the sake of completion, as it has been established to work well for opposing viewpoints but for relatively larger localization radii. Matching images from opposing viewpoints is challenging due to occlusion and lack of visual overlap~\cite{LOST}. This can be mitigated through local feature matching on orthographic views instead of perspective views. This results in localization within a significantly smaller radius of 7m, which compares to the use of a lower bound of 30m localization radius in LoST~\cite{LOST} due to lack of sufficient visual overlap in perspective views.
    
    The poor performance of deep descriptors like SuperGlue and D2-Net here can be attributed to a lack of rotation invariance. 

    Methods such as SIFT (Rectified Features)~\cite{RF}, although performing better than D2-Net~\cite{D2Net} and Superglue~\cite{superglue}, are worse than RoRD. We also compare the performance of our approach against Rectified Features~\cite{RF}, which is the conceptually closest approach to ours that does visual place recognition based on road surfaces. Due to the lack of open source code, we implement the orthographic view projection of road surfaces using our IPM approach as explained in section~\ref{IPM}. Following this, as mentioned in their work, we perform feature matching using the OpenCV implementation of SIFT~\cite{SIFT}. They use manual verification to decide if the retrieved database image in perspective view is a True Positive based on the visual overlap. This approach is susceptible to errors, thus we resort to the same automated evaluation protocol as used for other methods in this study.

\begin{table}
\caption{Quantitative results for the VPR task. We highlight \textbf{First} and \gray{Second} best recall values. We compare a multitude of methods for the VPR task on the Oxford RobotCar Dataset\cite{RobotCarDatasetIJRR}. All methods except LoST\cite{LOST} operate on proposed orthographic view (OV). Our method \textit{RoRD} outperforms other baseline methods significantly.}
\label{vpr_local}
\begin{center}
\scalebox{0.9}{
\begin{tabular}{cccccc}
\toprule

\textbf{Method} & LoST\cite{LOST} & D2-Net\cite{D2Net} & SP+SG\cite{superglue} &SIFT\cite{RF} & RoRD (Ours) \\
\midrule
\textbf{Recall} & 10.30 & 22.16 & 25.47 & \gray{34.97} & \textbf{70.31}\\
\bottomrule
\end{tabular}
}
\end{center}
\end{table}

\subsection{Qualitative Results}
We show qualitative results on three sequences for the \textit{DiverseView} dataset in Figure~\ref{fig:dv-qual}. These sequences comprise both indoor and outdoor scenes. Each row represents a particular sequence from the \textit{DiverseView} dataset. We conduct a three-way comparison between vanilla D2-Net, RoRD, and RoRD + OV (orthographic view correspondences projected back to perspective views). We observe that on all sequences, RoRD + OV and RoRD both outperform D2-Net. Even though the sequences contain rich features in the form of tables, laptops, graffiti walls and books, D2-Net fails to find precise correspondences.  When operating on orthographic views, our method (\textit{RoRD + OV}) finds even more correspondences, leading to reduced rotation and translation errors. 

Figure~\ref{fig:dv-qual2} and \ref{fig:dv-qual3} show qualitative results for the Oxford Robotcar~\cite{RobotCarDatasetIJRR} and the HPatches dataset~\cite{HPatches} respectively. For the task of finding precise pixel level correspondences between opposing viewpoint sequences, our method RoRD outperforms D2-Net where both the methods use orthographic views. As shown in Figure~\ref{fig:dv-qual3}, as a consequence of inplane-rotation training regime of \textit{RoRD}, we are able to obtain high quality feature correspondences and perform notably better than vanilla \textit{D2-Net}~\cite{D2Net}.

\section{CONCLUSION}

In this work, we present a system for creating \textit{rotation-robust features} to match local features between images under challenging viewpoint conditions. Our method via a training pipeline uses simple geometric operations such as in-plane rotations to achieve such features. Through these features, we show that we are able to obtain precise feature correspondences between image pairs despite significant viewpoint variations. We further extend the potential of such features by introducing an ensemble approach \textit{RoRD + CE} which combines the original information learnt from D2-Net\cite{D2Net} and the newly learnt \textit{rotation-robust features} (RoRD). On the HPatches dataset, we show that having this combination outperforms our other methods as well as existing baselines. For the Oxford RobotCar and the DiverseView Datasets, we further boost the performance of our method by using orthographic transformed images in lieu of perspective images. We show that orthographic views alone are not sufficient when used with existing local features, and rotation-robust descriptors are necessary to achieve superior performance as demonstrated on pose estimation and opposite-viewpoint place recognition tasks.

\bibliographystyle{IEEEtran}
\bibliography{IEEEexample}
\end{document}